\documentclass{article}
\usepackage{needspace}
\usepackage{hyperref} 
\usepackage[utf8]{inputenc}

\usepackage{spconf,amsmath,graphicx}

\usepackage{enumitem}
\setlist{nosep, leftmargin=14pt}

\usepackage{mwe} 

\title{Generative forecasting of brain activity enhances Alzheimer's classification and interpretation}

\name{Yutong Gao, Vince D. Calhoun, Robyn L. Miller}
\address{Tri-Institutional Center for Translational Research in Neuroimaging and Data Science (TReNDS)}
\begin{document}

\maketitle
\vspace{-15pt}
\begin{abstract}
Understanding the relationship between cognition and intrinsic brain activity through purely data-driven approaches remains a significant challenge in neuroscience. Resting-state functional magnetic resonance imaging (rs-fMRI) offers a non-invasive method to monitor regional neural activity, providing a rich and complex spatiotemporal data structure. Deep learning has shown promise in capturing these intricate representations. However, the limited availability of large datasets, especially for disease-specific groups such as Alzheimer’s Disease (AD), constrains the generalizability of deep learning models. In this study, we focus on multivariate time series forecasting of independent component networks derived from rs-fMRI as a form of data augmentation, using both a conventional LSTM-based model and the novel Transformer-based BrainLM model. We assess their utility in AD classification, demonstrating how generative forecasting enhances classification performance. Post-hoc interpretation of BrainLM reveals class-specific brain network sensitivities associated with AD.
\end{abstract}
\vspace{-5pt}


\footnotetext{Corresponding Author: Y.Gao Email: ygao11@gsu.edu}
\footnotetext{Research supported by NSF 2112455 and NIH R01AG073949}
\footnotetext{This work has been submitted to IEEE for possible publication. Copyright may be transferred, making this version potentially inaccessible.}

\vspace{-5pt}
\section{Introduction}
\vspace{-5pt}
\label{sec:intro}
Comprehending the relationship between cognition and intrinsic brain activity remains a significant challenge in neuroscience. Resting-state functional magnetic resonance imaging (rs-fMRI) provides a non-invasive approach to monitoring blood oxygen level fluctuations, providing insights into regional neural activity in the brain during task-free conditions. Given the high-dimensional and complex spatiotemporal dynamics of rs-fMRI data, deep learning models have shown promising capacity to learn rich representations, improving performance in classification and regression tasks.

Alzheimer’s Disease (AD) is a leading cause of dementia and the fifth leading cause of death among older adults in the U.S. \cite{alzheimer20182018}. Recognition of AD using purely data-driven approaches is crucial for public health interventions. However, training robust and generalizable deep learning models on rs-fMRI data is a significant challenge due to the limited availability of large datasets. Even large public datasets, such as the Alzheimer’s Disease Neuroimaging Initiative (ADNI), include relatively small patient groups, limiting model generalization at the group level. Data augmentation presents a practical solution to enhance dataset diversity without collecting additional data, with the goal of improving model performance.

Previous work has demonstrated the benefits of data augmentation through various techniques. For example, \cite{qiang2023functional} used a VAE-GAN framework to generate synthetic data for ADHD classification, showing improved results. Additionally, a multi-step recursive LSTM-based approach \cite{gao2024improving} was proposed to forecast future brain states for brain age prediction tasks. BrainLM \cite{ortega2023brainlm}, an encoder-decoder Transformer model, has demonstrated effectiveness in predicting future brain states. However, the model has yet to investigate how its predictions can be utilized for downstream tasks or how they might contribute to interpreting cognitive diseases.

In this study, we focus on multivariate time series forecasting of independent component networks (ICNs) derived from rs-fMRI, applying two deep learning-based approaches: Stateless LSTM and BrainLM. Our goal is to evaluate the efficiency of these models for forecasting future ICNs' time courses, and subsequently assess their utility for AD classification. We tackle the challenge of varying sequence lengths in the ADNI dataset by utilizing both truncation and replication techniques to ensure consistent input lengths for training deep learning models. Additionally, we conduct a thorough performance assessment to evaluate how the forecasting component contributes to the downstream classification task. Furthermore, post-hoc perturbation-based interpretation analysis on the trained BrainLM model reveals class-level sensitivity of brain networks related to forecasting future brain activity associated with AD.
\begin{figure*}[t]
    \centering
    \includegraphics[width=\textwidth]{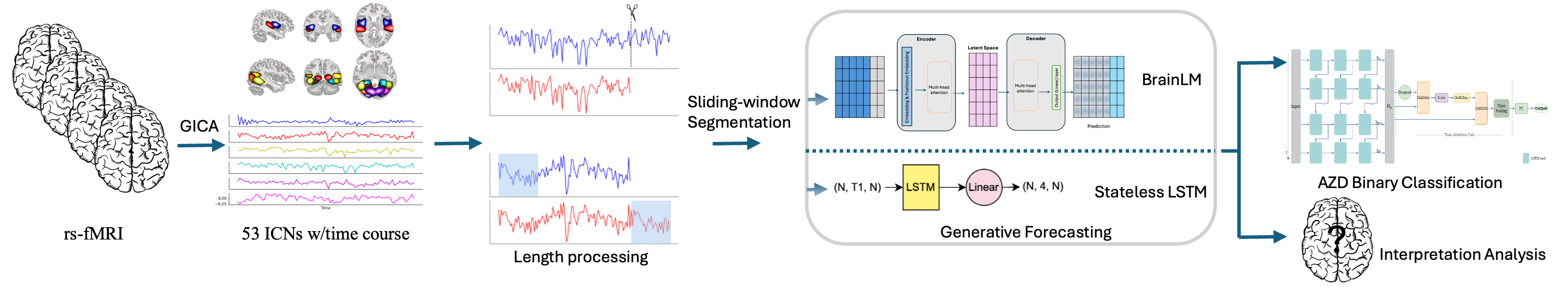}  
    \caption{The Generative Forecasting Pipeline for Classification and Interpretation. The rs-fMRI data is preprocessed and decomposed using GICA into 53 ICs with corresponding time courses. Extended scan time courses are truncated to create the baseline dataset, while regular scans with shorter lengths are replicated to align with the extended scans. The processed time series data are then segmented using a sliding window approach. Two generative forecasting models, stateless LSTM and BrainLM, are trained for forecasting. Data augmentation is performed using either the LSTM-based or BrainLM methods with replicated or truncated data. Finally, the TA-LSTM models are trained on the AD classification task to assess the effectiveness of the augmented data. A post-hoc perturbation-based interpretation analysis is conducted to examine class-level brain network sensitivity.}
    \label{fig:fullwidth}
    \vspace{-13pt} 
\end{figure*}

\vspace{-5pt}
\section{Methods}
\vspace{-5pt}
\subsection{Datasets and Preprocessing}
We conducted our research using the ADNI dataset (\url{http://adni.loni.usc.edu}), which includes data on healthy subjects and individuals at various stages of AD. For this study, we selected two groups based on the Clinical Dementia Rating (CDR): cognitively normal controls (CN, CDR = 0, 411 subjects) and Alzheimer’s patients (AD, CDR \( \geq 4 \), 95 subjects). Each subject had one rs-fMRI session included for analysis. The selected rs-fMRI scans were preprocessed using the Statistical Parametric Mapping software (SPM12). The preprocessing pipeline included rigid body correction, slice timing correction, and warping into an echo-planar imaging (EPI) template in standard Montreal Neurological Institute (MNI) space, with resampling to 3 × 3 × 3 mm³ voxels. We then applied group independent component analysis (GICA) to decompose the preprocessed rs-fMRI data into independent components (ICs) following the NeuroMark pipeline \cite{du2020neuromark}. Fifty-three ICs were selected across seven domains based on spatial location: subcortical (5), auditory (2), sensorimotor (9), visual (9), cognitive control (17), default mode (7), and cerebellar (4), with the number of ICs in each domain indicated in parentheses. A detailed table of these components can be found in \cite{gao2023interpretable}. 

The ADNI dataset consists both regular (137-timestamp) and extended (194-timestamp) scans, requiring adjustments to support deep learning model training. We addressed this by both truncating the longer sequences and replicating the shorter ones to standardize input length across all sequences. The illustration is presented in Fig. 1.

\label{sec:format}
\vspace{-5pt}
\subsection{Generative Forecasting Models}
\textbf{Brain Language Model (BrainLM)} BrainLM was initially introduced by \cite{ortega2023brainlm}. For this study, we modified BrainLM by removing the spatial input and its corresponding embedding layer to better suit our research objectives. The model architecture, shown in Fig. 2, follows a transformer-based encoder-decoder structure, with multi-head attention layers forming the core of the transformer. In our adaptation, we masked the final 1/6 of time points, which were subsequently encoded and decoded for reconstruction.

\textbf{Stateless LSTM} is a conventional model architecture that leverages inherent sequence-prediction capabilities, featuring input, output, and forget gates along with cell states, to forecast subsequent time points. This model comprises a single LSTM layer with a hidden size of 50, followed by a linear output layer that generates predictions for four consecutive time points for each network, as shown in Fig. 1.
\vspace{-5pt}
\subsection{Classification Model}
The TA-LSTM model, introduced in \cite{gao2023interpretable}, was utilized to evaluate the original, replicated, and augmented datasets for binary classification between CN and AD groups. This model comprises three LSTM layers, each with 64 hidden units, followed by a time-attention layer and an output layer. The time-attention layer applies scaled dot-product attention, reducing attention scores to a single value at each time point. These scores are then passed through a fully connected layer to generate class probability scores.

\begin{figure*}[htbp]
    \centering
    \includegraphics[width=\textwidth]{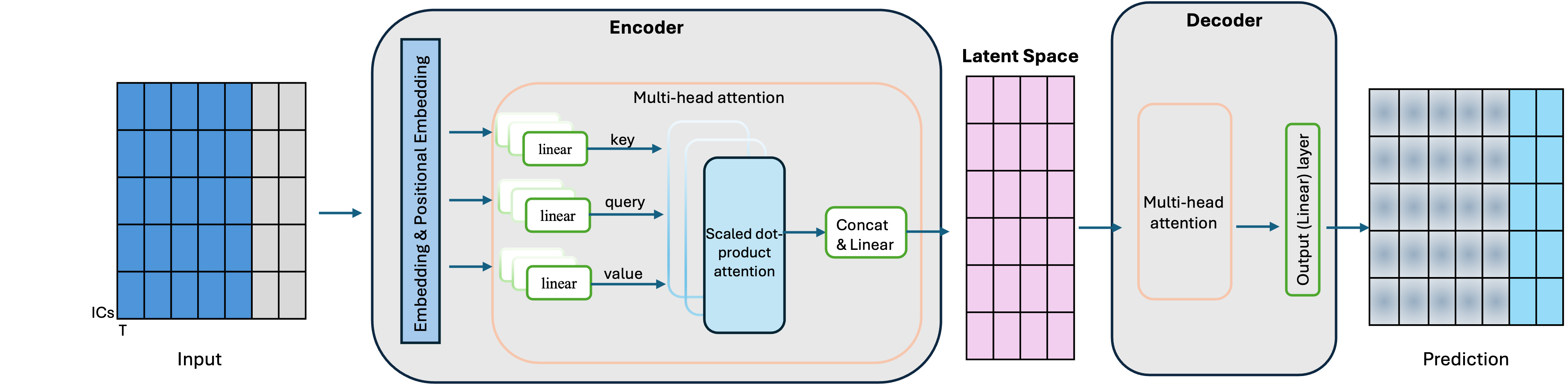}  
    \caption{BrainLM model architecture}
    \label{fig:fullwidth}
    \vspace{-13pt} 
\end{figure*}

\label{sec:format}
\vspace{-5pt}
\section{Experiments and Results}
\vspace{-5pt}
\label{sec:pagestyle}
\subsection{Experiment setting}

For each preprocessed fMRI independent component (IC) time courses, the extended scans of 194 timestamps were truncated to 137 timestamps, which served as the baseline. Replication, a straightforward oversampling data augmentation technique, was also evaluated. The 137 timestamps were concatenated with the first 57 replicated timestamps to form a 194-timestamp sequence. The time series were segmented into 24-timestep windows using a sliding window approach with a step size of 4. The initial 20 time points served as the training data for forecasting the subsequent 4 time points, which acted as labels for the stateless LSTM model. For BrainLM predictions, the last 1/6 time points within each window were masked, as BrainLM is trained to make predictions through reconstruction for forecasting. Two generative forecasting models were trained on 80\% of the segmented time series data and evaluated on the remaining 20\%. The models were trained with a batch size of 32 using the Adam optimizer for 500 epochs, minimizing the mean squared error between the predicted and actual time series.

After training, the last 20 timestamps of 137- or 194-timestamp were passed into the trained forecasting models to forecast the subsequent 4 unseen timestamps. For the BrainLM model, 4 random timestamps were concatenated to form a 24-timestamp input for forecasting via reconstruction. Six time series datasets were evaluated for the downstream AD classification task: (a) baseline (t=137), (b) baseline with Stateless LSTM (t=141), (c) baseline with BrainLM (t=141), (d) baseline with replication (t=194), (e) replication with Stateless LSTM (t=198), and (f) replication with BrainLM (t=198). The 90\% datasets were trained using a five-fold cross-validation scheme, and 10\% for held-out testing. The split was performed using a stratified group approach, ensuring that the class ratios in the test set matched those in the full dataset. The train-test process was repeated five times with different random seeds, resulting in 25 test scores, providing a robust performance assessment. The models were trained with a batch size of 32 using the Adam optimizer for 800 epochs. The best-performing models on the validation set were used for testing on the held-out test set.
\begin{figure}[htb]

  \centering
  \centerline{\includegraphics[width=8.5cm]{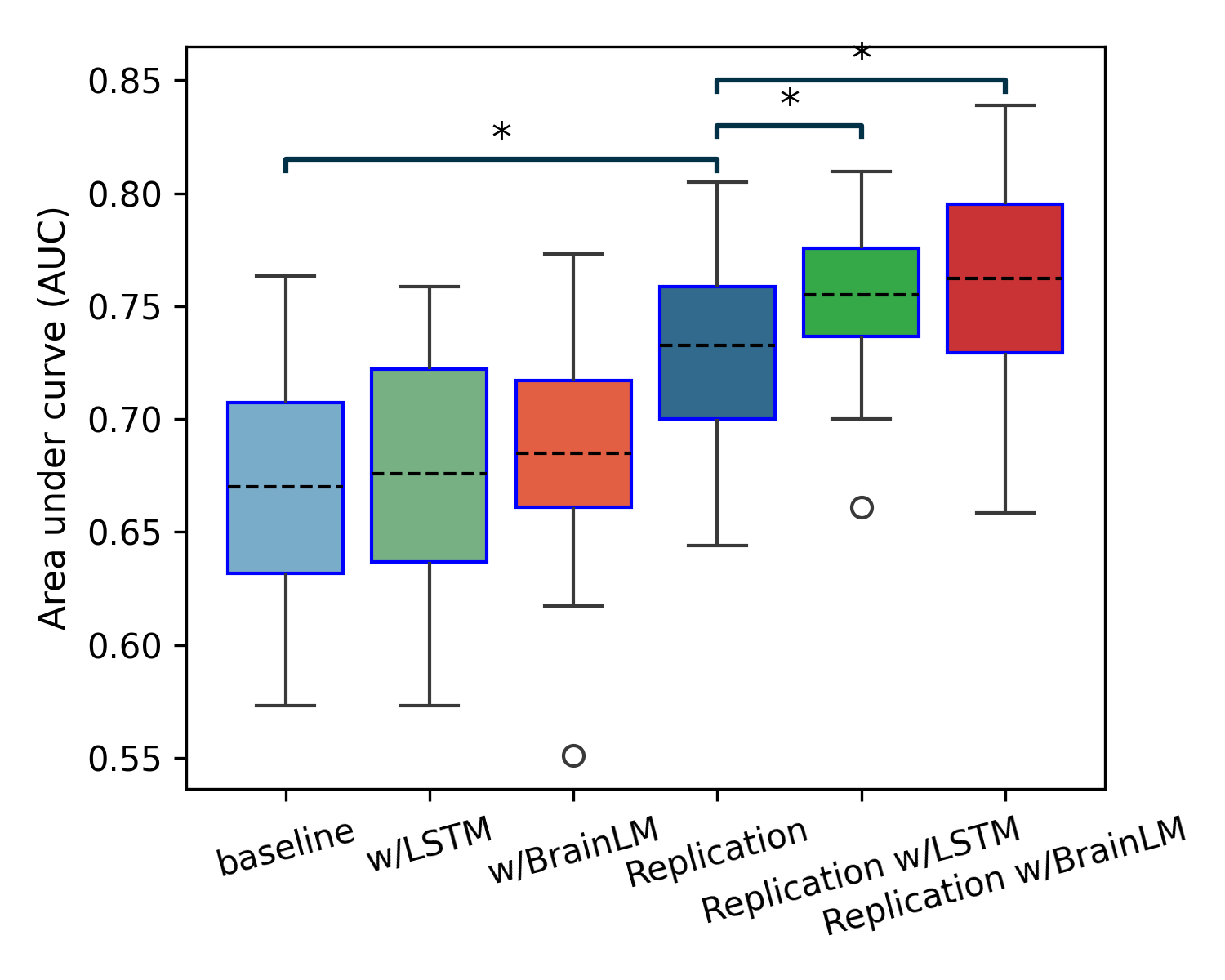}}
  \caption{TA-LSTM AD Classification Result.}
  \vspace{-13pt} 
\end{figure}
\vspace{-5pt}
\subsection{AD Classification Result}
The binary classification of AD and CN was evaluated using the Area Under the Curve (AUC) metric. Fig. 3 presents the results depicting the AUC scores across different experimental conditions: baseline, baseline w/stateless LSTM, baseline w/BrainLM, baseline w/replication, replication w/stateless LSTM, replication w/BrainLM. The dashed black lines show the mean AUC of 25 tests.

The results indicate clear trends in model performance improvement with replication and generative forecasting model augmentation. The baseline model achieved an average AUC of 0.67, while the addition of Stateless LSTM forecasting resulted in an AUC of 0.676, and BrainLM prediction improved it to 0.685. This suggests that incorporating generative modeling techniques enhances data representation learning for better classification performance. Notably, replicating shorter time series yields a performance gain of 0.732, indicating that TA-LSTM classification further benefits from increased input length. It is important to note that the performance improvement is not solely due to simple replication; rather, the replication also provides extra information from the longer duration of extended scans. The models utilizing the replicated data for training in forecasting augmentation achieved an AUC of 0.755 with Stateless LSTM and 0.762 with BrainLM. Both results show a statistically significant difference (p \( \leq 0.05 \)) compared to the baseline w/replication, suggesting that augmentation with longer scans further enhances performance. Overall, replication with BrainLM achieves the highest mean AUC among all experimental settings.

\vspace{-5pt}
\subsection{Brain Interpretation}
Perturbation-based analysis of the trained BrainLM model identified the most sensitive brain regions during forecasting. For each IC, we set its activity to zero and evaluated the BrainLM model separately for CN and AD classes. Sensitivity was assessed by calculating the percentage change in loss (delta) upon silencing each IC.

The five most sensitive networks for each class are presented in Fig. 4. The red regions indicate that the AD group exhibits greater sensitivity than the CN group, reflecting the effects of silencing the network. In AD, certain networks are often deactivated, disrupted, or dysregulated due to disease progression, resulting in suboptimal functioning and weakened connectivity. When these compromised networks are perturbed, high sensitivity typically reflects dysregulation, highlighting the network’s instability. These findings are consistent with existing work; for example, the parahippocampus is an important region previously implicated in AD. Conversely, blue regions indicate that the CN group shows greater sensitivity than the AD group. In CN, networks usually operate with stable and robust connectivity. When a network in this group is perturbed, high sensitivity often signifies active engagement, as the network’s stability and functionality enable it to respond dynamically. This observation is also consistent with previous studies indicating that CN generally has more active activity observed in the anterior cingulate cortex (ACC) in the default mode compared with AD. These findings suggest that both types of sensitivity could serve as potential biomarkers for differentiating between CN and AD.

\begin{figure}[htb]
  \centering
  \centerline{\includegraphics[width=8.5cm]{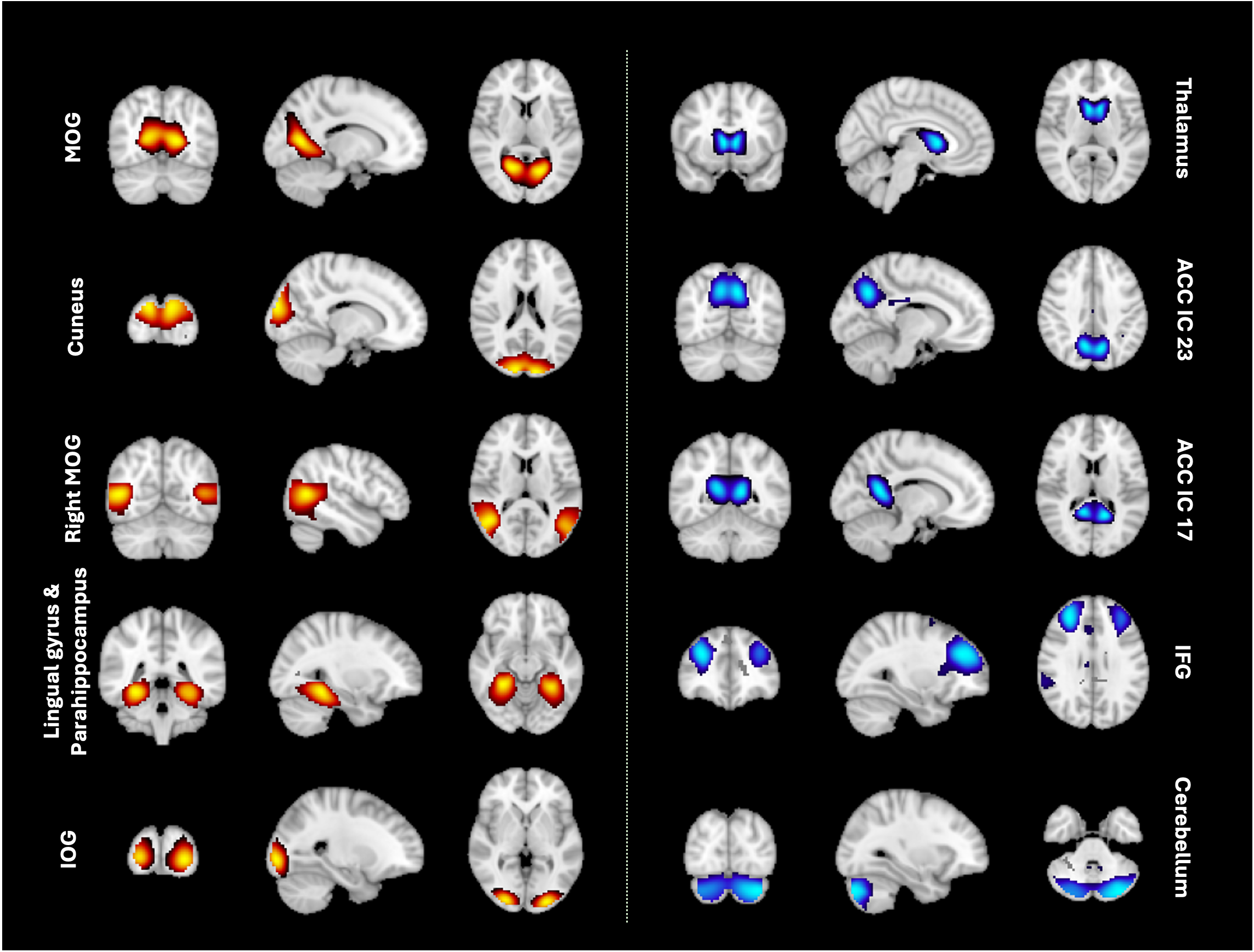}}
  \caption{ Perturbation-based CN-AD Brain Network Sensitivity Interpretation. Sensitivity levels: Red: AD \textgreater CN; Blue: CN \textgreater AD. (MOG: Middle Occipital Gyrus; IOG: Inferior Occipital Gyrus; ACC: Anterior Cingulate Cortex; IFG: Inferior Frontal Gyrus)}
  \vspace{-15pt} 

\end{figure}
\vspace{-5pt}

\section{Concolusions}
In this study, we address data limitations in deep learning training for AD classification by applying both a conventional LSTM-based model and the novel BrainLM model for generative forecasting, assessing their utility in AD classification. We address the challenge of varying sequence lengths across clinical scans by using truncation and replication to standardize input lengths, suggesting that simple replication, rather than cropping, is a more effective approach for handling length invariance. Our findings indicate that generating longer time courses through deep learning-based forecasting enhances the accuracy of AD classification task, with BrainLM showing greater improvement compared to stateless-LSTM. Additionally, a post-hoc analysis was conducted to interpret the BrainLM model by assessing the sensitivity of forecasting accuracy with respect to individual brain networks, providing interpretability insights.
\label{sec:Discussion}
\vspace{-5pt}

\section{Compliance with Ethical Standards}
This study utilized human subject data from the ADNI dataset (\url{http://adni.loni.usc.edu}), which received ethical approval from the Institutional Review Board (IRB). No additional ethical review was required, as confirmed by the dataset’s open-access license.

\bibliographystyle{IEEEbib}
\bibliography{main}

\begin{thebibliography}{1}

\bibitem{alzheimer20182018}
Alzheimer's Association et~al.,
\newblock ``2018 alzheimer's disease facts and figures,''
\newblock {\em Alzheimer's \& Dementia}, vol. 14, no. 3, pp. 367--429, 2018.

\bibitem{qiang2023functional}
Ning Qiang, Jie Gao, Qinglin Dong, Huiji Yue, Hongtao Liang, Lili Liu, Jingjing Yu, Jing Hu, Shu Zhang, Bao Ge, et~al.,
\newblock ``Functional brain network identification and fmri augmentation using a vae-gan framework,''
\newblock {\em Computers in Biology and Medicine}, vol. 165, pp. 107395, 2023.

\bibitem{gao2024improving}
Yutong Gao, Charles~A Ellis, Vince~D Calhoun, and Robyn~L Miller,
\newblock ``Improving age prediction: Utilizing lstm-based dynamic forecasting for data augmentation in multivariate time series analysis,''
\newblock in {\em 2024 IEEE Southwest Symposium on Image Analysis and Interpretation (SSIAI)}. IEEE, 2024, pp. 125--128.

\bibitem{ortega2023brainlm}
Josue Ortega~Caro, Antonio~Henrique Oliveira~Fonseca, Christopher Averill, Syed~A Rizvi, Matteo Rosati, James~L Cross, Prateek Mittal, Emanuele Zappala, Daniel Levine, Rahul~M Dhodapkar, et~al.,
\newblock ``Brainlm: A foundation model for brain activity recordings,''
\newblock {\em bioRxiv}, pp. 2023--09, 2023.

\bibitem{du2020neuromark}
Yuhui Du, Zening Fu, Jing Sui, Shuang Gao, Ying Xing, Dongdong Lin, Mustafa Salman, Anees Abrol, Md~Abdur Rahaman, Jiayu Chen, et~al.,
\newblock ``Neuromark: An automated and adaptive ica based pipeline to identify reproducible fmri markers of brain disorders,''
\newblock {\em NeuroImage: Clinical}, vol. 28, pp. 102375, 2020.

\bibitem{gao2023interpretable}
Yutong Gao, Noah Lewis, Vince~D Calhoun, and Robyn~L Miller,
\newblock ``Interpretable lstm model reveals transiently-realized patterns of dynamic brain connectivity that predict patient deterioration or recovery from very mild cognitive impairment,''
\newblock {\em Computers in Biology and Medicine}, vol. 161, pp. 107005, 2023.

\end{thebibliography}

\end{document}